\documentclass[11pt,a4paper]{article}
\usepackage{times,latexsym}
\usepackage{url}
\usepackage[T1]{fontenc}
\usepackage{graphicx}
\usepackage{adjustbox}
\usepackage{amssymb}
\usepackage[acronym]{glossaries}
\usepackage[table]{xcolor}
\usepackage{longtable}

    \usepackage[acceptedWithA]{tacl2021v1}
\usepackage{xspace,mfirstuc,tabulary}

\newif\iftaclinstructions
\taclinstructionsfalse % AUTHORS: do NOT set this to true
\iftaclinstructions
\renewcommand{\confidential}{}
\renewcommand{\anonsubtext}{(No author info supplied here, for consistency with
TACL-submission anonymization requirements)}
\newcommand{\instr}
\fi

\iftaclpubformat % this "if" is set by the choice of options

\else

\fi

\title{AfriWOZ: Corpus for Exploiting Cross-Lingual Transferability for Generation of Dialogues in Low-Resource, African Languages}
 \author{\\
   Tosin Adewumi\Thanks{corresponding author}$^\diamond$$^1$$^\dagger$,
   Mofetoluwa Adeyemi$^\dagger$,
   Aremu Anuoluwapo$^\dagger$,
   Bukola Peters$^2$,
   Happy Buzaaba$^\dagger$, \\
   Oyerinde Samuel$^\dagger$,
   Amina Mardiyyah Rufai$^\dagger$,
   Benjamin Ajibade$^\dagger$,
   Tajudeen Gwadabe$^\dagger$, \\
   Mory Moussou Koulibaly Traore$^\dagger$,
   Tunde Ajayi$^\dagger$,
   Shamsuddeen Muhammad$^\dagger$,
   Ahmed Baruwa$^\dagger$, \\
   Paul Owoicho$^\dagger$,
   Tolulope Ogunremi$^\dagger$,
   Phylis Ngigi$^\dagger$,
   Orevaoghene Ahia$^\dagger$,
   Ruqayya Nasir$^\dagger$, \\
   Foteini Liwicki$^1$
   \and
   Marcus Liwicki$^1$
   \\
   $^1$ML Group, Luleå University of Technology/ $^\dagger$Masakhane/ $^2$CIS
 }

\date{}

\makeglossaries

\newacronym{nlp}{NLP}{Natural Language Processing}
\newacronym{ner}{NER}{Named Entity Recognition}
\newacronym{sa}{SA}{Sentiment Analysis}
\newacronym{bow}{BoW}{bag-of-words}
\newacronym{cbow}{CBoW}{continuous Bag-of-Words}
\newacronym{sltc}{SLTC}{Swedish Language Technology Conference}
\newacronym{ann}{ANN}{artificial neural network}
\newacronym{nn}{NN}{neural network}
\newacronym{lstm}{LSTM}{Long Short Term Memory Network}
\newacronym{sota}{SoTA}{state-of-the-art}
\newacronym{nlg}{NLG}{Natural Language Generation}
\newacronym{mwe}{MWE}{Multi-Word Expression}
\newacronym{sw}{SW}{Simple Wiki}
\newacronym{mt}{MT}{Machine Translation}
\newacronym{gdc}{GDC}{Gothenburg Dialog Corpus}
\newacronym{nlu}{NLU}{Natural Language Understanding}
\newacronym{ai}{AI}{Artificial Intelligence}
\newacronym{iaa}{IAA}{inter-annotator agreement}
\newacronym{multiwoz}{MultiWOZ}{Multi-Domain Wizard-of-Oz}
\newacronym{dialogpt}{DialoGPT}{dialogue generative pre-trained transformer}
\newacronym{bpe}{BPE}{byte-pair encoding}
\newacronym{cus}{CUS}{credibility unanimous score}
\newacronym{bleu}{BLEU}{bilingual evaluation understudy}
\newacronym{bert}{BERT}{Bidirectional Encoder Representations from Transformers}
\newacronym{t5}{T5}{Text-to-Text Transfer Transformer}
\newacronym{bart}{BART}{Bidirectional and Auto-Regressive Transformers}
\newacronym{xlmr}{XLM-R}{Cross-Lingual Model-RoBERTa}
\newacronym{asr}{ASR}{automatic speech recognition}
\newacronym{pii}{PII}{personally identifiable information}

\begin{document}
\maketitle
% \vspace*{20px}
\begin{abstract}
Dialogue generation is an important \acrshort{nlp} task fraught with many challenges.
The challenges become more daunting for low-resource African languages.
% , in terms of data.
%Generation of dialogues is known to be a challenging task for many reasons.
%It becomes more challenging for African languages which are low-resource in terms of data.
To enable the creation of dialogue agents for African languages, we contribute the first high-quality dialogue datasets for 6 African languages: Swahili, Wolof, Hausa, Nigerian Pidgin English, Kinyarwanda \& Yor\`{u}b\'{a}.
These datasets consist of 1,500 turns each, which we translate from a portion of the English multi-domain MultiWOZ dataset.
%Subsequently, we investigate how to create high-quality dialogue models for African languages through cross-lingual transfer.
Subsequently, we investigate \& analyze the effectiveness of modelling
%starting with pre-trained models and
through transfer learning
%for these African languages
by utilziing \acrfull{sota} deep monolingual models: \acrshort{dialogpt} and BlenderBot.
We compare the models with a simple seq2seq baseline using perplexity.
%an automatic metric.
Besides this, we conduct human evaluation of single-turn conversations by using majority votes and measure \acrfull{iaa}.
We find that the hypothesis that deep monolingual models learn some abstractions that generalize across languages holds.
We observe human-like conversations, to different degrees, in 5 out of the 6 languages.
%It, however, applies to different degrees in different languages, which is expected.
The language with the most transferable properties is the Nigerian Pidgin English, with a human-likeness score of 78.1\%, of which 34.4\% are unanimous.
% The main contributions of this paper include the representation (through the provision of high-quality dialogue data) of under-represented African languages and demonstrating the cross-lingual transferability hypothesis for dialogue systems.
We freely provide the datasets and host the model checkpoints/demos on the HuggingFace hub for public access.

%investigate the possibility of cross-lingual transfer from a \acrfull{sota} deep monolingual model (\acrshort{dialogpt}) to 6 African languages and compare with 2 baselines (BlenderBot 90M, another \acrshort{sota}, and a simple Seq2Seq).
%Therefore, we translate a small portion of the English multi-domain MultiWOZ dataset for each target language.
%Besides intrinsic evaluation (i.e. perplexity), we conduct human evaluation of single-turn conversations by using majority votes and measure \acrfull{iaa}.
%using Fleiss Kappa and credibility tests.
%The results show that the hypothesis that deep monolingual models learn some abstractions that generalise across languages holds.
%We observe human-like conversations in 5 out of the 6 languages. It, however, applies to different degrees in different languages, which is expected.
%The language with the most transferable properties is the Nigerian Pidgin English, with a human-likeness score of 78.1\%, of which 34.4\% are unanimous.
%Its credibility \acrshort{iaa} unanimous score is 66.7\%.

%\\
%\textbf{Keywords: }Chatbots, African Languages, Conversational Systems, Open Domain, Low-resource
\end{abstract}

% \vspace*{20px}

\section{Introduction}
The ability to understand and converse fluently in natural language is considered a major component of intelligence.
% Hence, the degree of intelligence of an \acrfull{ai} system is reflected in its \acrfull{nlg} capabilities.
Over the years, open-domain conversational (or dialogue) systems have evolved \cite{weizenbaum1969computer,zhang2020dialogpt,roller-etal-2021-recipes,adiwardana2020towards,adewumi2019conversational}.
Advances in deep neural networks, such as the Transformer-based architectures, have brought improvements to the field \cite{vaswani2017attention,devlin2018bert,radford2019language,he2020deberta}.
These models have demonstrated \acrshort{sota} performances in \acrfull{nlu} and \acrfull{nlg} tasks \cite{wang2019superglue,gehrmann2021gem}.

While significant advancements have been made in the field, the majority of focus has been on the English language.
For example, many models were originally pretrained on English data, though researchers have recently been producing some multilingual versions \cite{devlin2018mbert,NEURIPS2019_c04c19c2,xue-etal-2021-mt5}.
Some of these multilingual models, however, have been shown to have poor performance compared to models trained completely on the target language data \cite{virtanen2019multilingual,ronnqvist2019multilingual}.
\acrshort{nlp} challenges get more daunting for languages that do not have sufficient data to train with, usually called low-resource languages \cite{nekoto-etal-2020-participatory,adewumi2020challenge,10.1162/tacl_a_00416}.
Thus, the multilingual versions of the deep models do not cover many of these languages.
For example, Table \ref{langrep} reveals languages not covered by some of the multilingual models and Google Machine Translate\footnote{As of April 15, 2022}.
This shows many languages are still under-represented.
Besides the challenge of availability of data or high-quality data, there are also technical \cite{roller-etal-2021-recipes} and ethical challenges \cite{dinan-etal-2020-queens,javed2021understanding}.

%The advancements notwithstanding, challenges still exist with building open-domain conversational systems \cite{jurafsky2020speech, zhang2020dialogpt}.
%These challenges include technical \cite{roller-etal-2021-recipes} and ethical challenges \cite{javed2021understanding}.
%This is more so that many of the models are 

%originally pretrained on English data, though researchers have recently been producing multilingual versions of some of the models \cite{ogueji-etal-2021-small,devlin2018mbert,NEURIPS2019_c04c19c2,reid-etal-2021-afromt,xue-etal-2021-mt5}.

\begin{table}[h]
\centering
\resizebox{\columnwidth}{!}{%
\begin{tabular}{lcccccc}
\hline
\textbf{Language} &
\multicolumn{6}{c}{\textbf{Multilingual model}}
\\
%\hline
 & m\acrshort{bert} & m\acrshort{bart} & m\acrshort{t5}  & \acrshort{xlmr} & AfriBERTa & Google \acrshort{mt}
 % Phylum, Region, Speakers, ISO code
\\
\hline
Pidgin English & X & X & X & X & $\sqrt{}$ & X
\\
Yor\`{u}b\'{a} & $\sqrt{}$ & X & $\sqrt{}$ & X & $\sqrt{}$ & $\sqrt{}$
\\ % \hline
Hausa & X & X & $\sqrt{}$ & $\sqrt{}$ & $\sqrt{}$ & $\sqrt{}$
\\ % \hline
Wolof & X & X & X & X & X & X
\\
Swahili & $\sqrt{}$ & X & $\sqrt{}$ & $\sqrt{}$ & $\sqrt{}$ & $\sqrt{}$
\\
Kinyarwanda & X & X & X & X & X & $\sqrt{}$
\\
\hline
\end{tabular}
}
\caption{\label{langrep} The languages in some models: \footnotesize{$\sqrt{}$: yes, X: no}}
\end{table}

%Since many of the multilingual models are not autoregressive models and pretraining (say, a Transformer) in low-resource languages requires having large data (or preferrably conversational data).
%Hence the motivation to use English pre-trained models and fine-tune them on the target languages.\acrshort{nlg} in open-domain conversational systems aims to generate text that engage users in meaningful conversation.

The motivation and contributions of this study are to (1) create the first high-quality dialogue datasets for the target languages from the benchmark MulitWOZ dataset \cite{budzianowski-etal-2018-multiwoz} and (2) investigate by transfer learning, for open-domain dialogue systems, the hypothesis that  deep  monolingual  models  learn some abstractions that generalise across languages \cite{artetxe-etal-2020-cross}.
The contribution of the data provides other side-contributions because they may be adapted for other \acrshort{nlp} tasks, such as \acrfull{mt}, task-based dialogue systems and \acrfull{asr}, among others.
We obtain promising results that apparently validate the stated hypothesis and obtain better human evaluation results for 2 of the languages than what was shown for Swedish in a similar setup by \citet{adewumi2021sm}.
We freely provide the codes, datasets\footnote{github.com/masakhane-io/chatbots-african-languages}
and model checkpoints/demos for public use on the HuggingFace hub\footnote{huggingface.co/tosin}.
The findings of this study seem to be the first for the languages concerned under open-domain dialogue setting, to the best of our knowledge.

%The language most transferable to appears to be Nigerian Pidgin English.
%78.1\% of its conversations are judged as human-like, which is more than 20\% higher than that obtained by \citet{adewumi2021sm}.

%We present results using perplexity \cite{brown1992estimate} and human evaluation.
%We measure \acrshort{iaa} scores using credibility tests and Fleiss Kappa \cite{fleiss1971measuring}.

%The paper is short of a clear motivation/scientific question and new insights/findings to the community.
% - Lack of a clear scientific or engineering question that the paper aims to evaluate.
%Is it possible for a model trained in one language to `speak' other different languages after finetuning on little data?
%We show that this is possible.

The rest of this paper is organised as follows.
The `languages of the study' section (\ref{sec:los}) presents brief details of the languages; the methodology section (\ref{sec:method}) describes the experimental setup, data models and modes of evaluation, including the newly-introduced \acrfull{cus} for \acrshort{iaa}; the results \& discussion section (\ref{sec:res}) presents the tables of results and evaluation for all the models, including the error analysis; the conclusion section (\ref{sec:conc}) then follows after the related work section (\ref{sec:related}) and limitation section (\ref{sec:limits}).

\section{Languages of the Study}
\label{sec:los}

The vast majority of work in dialogue focuses on high-resource languages like English \cite{zhang2020dialogpt,adiwardana2020towards}, Chinese, with limited work in other high-resource languages.
We focus on 6 African languages spoken by millions of people.
The languages were selected for this work based on their diversity and the availability of contributors.
They cover countries in West, East, Central and Southern Africa \cite{heine2000african} and over 239 million speakers combined.
Examples of translated sentences for each language are given in Table \ref{sampleconv}.
The examples are from the training set of the English MultiWOZ dataset.

% \cellcolor[HTML]{BBACBB}
% \cellcolor[HTML]{AA88FF}

\begin{table*}[h]
\small
\centering
%\resizebox{\columnwidth}{!}{%
%\begin{tabular}{l|c|c}
\begin{tabular}{p{.23\textwidth}p{.77\textwidth}}
\hline
\textbf{Language} & \textbf{Example of 3 Conversation turns}
\\
\hline
English & 
 I have several options for you; %do you prefer African, Asian, or British food?
\\
 &  I want to book it for 2 people and 2 nights starting from Saturday.\\
 &  That is all I need to know. Thanks, good bye.
\\
\hline
Nigerian Pidgin English & 
 I get plenty options for you;% you prefer African, Asian, or British food?
\\
 & I wan book am for 2 people for 2 night for Saturday\\
 &  Na everything wey i need to know. thank you. good bye
\\
\hline
Yor\`{u}b\'{a} &  Mo n\'{i} aw\d{\'{o}}n \`{a}\d{s}\`{a}y\`{a}n p\'{u}p\d{\`{o}} f\'{u}n \d{o}; %\d{s}\'{e} o f\d{\'{e}}r\`{a}n \'{o}unj\d{e} \'{A}fr\'{i}k\`{a}, \'{A}s\'{i}\`{a}, t\`{a}b\'{i} \`{i}l\'{u} G\d{\`{e}}\d{\'{e}}s\`{i}?
\\
 & Mo f\'{e} \d{s}e \`{i}w\'{e} f\'{u}n \`{e}n\`{i}y\`{a}n m\'{e}j\`{i} \`{a}ti f\'{u}n al\d{\'{e}} m\'{e}j\`{i} t\'{i} \'{o} b\d{\'{e}}r\d{\`{e}} l\'{a}ti \d{o}j\d{\'{o}} S\'{a}t\`{i}de\'{e}.\\
 &  \`{I}y\d{e}n ni gbogbo ohun t\'{i} mo n\'{i}l\`{o} l\'{a}ti m\d{\`{o}}. O \d{s}eun, \'{O} d\`{a}b\`{o}.
\\
\hline
Hausa & 
 Ina da zabubbuka da yawa a gare ku;% kun fi son abincin Afirka, Asiya, ko Biritaniya?
\\
 & Ina so in yi wa mutane 2 da dare 2 farawa daga ranar Asabar.\\
 & Wannan shine kawai abin da nake bukatar sani. Godiya, bye bye.
\\
\hline
Wolof & 
 Amna ay tanneef yu bari ngir yaw; %ndax bëg ngan lekku niit ñu ñull yi, wa asi wala wa angalteer?
\\
 & Soxla jënd ngir ñaari niit ak ñaari guddi mu tambelee Gawu\\
 & Dedet  li rek la soxla. jerejef. ba benen yoon
\\
\hline
Swahili & Nina chaguzi kadhaa kwako;% unapendelea chakula cha Kiafrika, Kiasia, au Uingereza?
\\
 & Nataka kuihifadhi kwa watu 2 na usiku 2 kuanzia Jumamosi.\\
 & Hiyo ndiyo yote ninahitaji kujua. Asante, kwaheri.
\\
\hline
Kinyarwanda & 
 Mfite henshi naguhitiramo hari; %ibiryo bitetse mu buryo bw' Afrika, Aziya, cyangwa Ubwongereza?
\\
 & Ndashaka kubika imyanya ku bantu 2 n'amajoro 2 guhera ku wa Gatandatu.\\
 & Ibyo ni byo nari nkeneye kumenya. Urakoze, murabeho.
\\
\hline
\end{tabular}
%}
\caption{\label{sampleconv} Translation examples from the English MultiWOZ data for the six languages.}
%}
\end{table*}

\paragraph{Swahili}
Swahili is a Bantu language.
It is spoken by the Bantu people in the southern half of Africa \cite{polome1967swahili}.
It is an official language of the East African Community (EAC) countries.
These include: Uganda, Burundi, Kenya, Tanzania, Rwanda, South Sudan and the Democratic Republic of the Congo (DRC).
It is a lingua franca of other areas like Malawi, Mozambique, the southern tip of Somalia, and Zambia \cite{polome1967swahili}.
There are more than 50 million speakers of the language. \footnote{swahililanguage.stanford.edu}
It is also one of the working languages of the African Union.

\paragraph{Wolof}
Wolof is spoken in Senegal, Mauritania and the Gambia.
More than 7 million people are believed to speak the language\footnote{worlddata.info/languages/wolof.php}.
It is of the Senegambian branch of the Niger–Congo language phylum, which is the largest language phylum in the world \cite{heine2000african}.
Unlike most other languages of the Niger-Congo phylum, Wolof is not a tonal language.

\paragraph{Hausa}
Hausa is a Chadic language spoken by the Hausa people.
It is mainly within the northern part of Nigeria and the southern part of Niger.
It has significant minorities in Cameroon, Chad, and Benin.
It is the most widely spoken language within the Chadic branch of the Afroasiatic phylum \cite{heine2000african}.
It has more than 40 million speakers\footnote{britannica.com/topic/Hausa-language}.

\paragraph{Nigerian Pidgin English}
Nigerian Pidgin English is a grammatically simplified means of communication among the ethnic groups in Nigeria.
Its vocabulary and grammar are limited and often drawn from the English language.
It is popular among young people \cite{ihemere2006basic}.
About 75 million are estimated to speak the language but the exact number is difficult to estimate since it is not an official language\footnote{bbc.com/news/world-africa-38000387}.

\paragraph{Kinyarwanda}
Kinyarwanda is an official language of Rwanda and a dialect of the Rwanda-Rundi language spoken in Rwanda \cite{heine2000african}.
It is one of the four official languages of Rwanda.
Over 22 million people are estimated to speak the language\footnote{worlddata.info/languages/kinyarwanda.php}.

\paragraph{Yor\`{u}b\'{a}}
Yor\`{u}b\'{a} is predominantly spoken in Southwestern Nigeria by the ethnic Yor\`{u}b\'{a} people \cite{heine2000african}.
It is primarily spoken in a dialectal area spanning Nigeria and Benin with smaller migrated communities in Cote d'Ivoire, Sierra Leone and The Gambia.
The number of Yor\`{u}b\'{a} speakers is more than 45 million\footnote{worlddata.info/languages/yoruba.php}.

\section{An African Dialogue Dataset: AfriWOZ}
The Yor\`{u}b\'{a} language has small dialogue data online\footnote{YorubaYeMi-textbook.pdf \& theyorubablog.com}, unlike the other languages.
We chose to use these sources for Yor\`{u}b\'{a} because of the local entities represented in the data and then augment the data if necessary.
As a result of the scarcity or non-existent dialogue data for most of the languages, the authors
decided to translate an English dialogue dataset.
The poll was between Reddit\footnote{reddit.com/} and MultiWOZ \cite{budzianowski-etal-2018-multiwoz}.
Most contributors voted in favour of MultiWOZ, though it is from task-oriented dialogues, instead of Reddit because of the high probability of toxic content \cite{roller-etal-2021-recipes}.
Indeed, in order to address the challenge of toxic comments in dialogues \cite{dinan2019build}, \citet{solaimanprocess} advocated for the approach of carefully curating dataset as a safe approach.
They observed that the adjustment of a model's behavior is possible with a small, hand-curated dataset.
This approach takes ethical considerations into account \cite{jurafsky2020speech,javed2021understanding}.
We follow this approach.

\subsection{\acrshort{multiwoz}}
\acrshort{multiwoz} is a collection of human-human written conversations that span multiple domains and topics.
It has gone through improvements and extensions over the years and currently has about 10,000 dialogues \cite{eric-EtAl:2020:LREC}.
Its multiple domain/topic coverage, though limited, makes it ideal for open-domain modeling.
Indeed, \citet{budzianowski-etal-2018-multiwoz} experimented with it for neural response generation, showing its usefulness across a range of dialogue tasks.
It has over 113,000 turns in the training set and over 14,700 turns each in both the validation and test sets.
Some of the domains covered are hospital, police, attraction, hotel, restaurant, taxi, train and booking.

In our work, we extracted and translated the first 1,000 turns from the training set and the first 250 turns each from the validation and test sets for the languages.
Only 200 turns from the \acrshort{multiwoz} training set were added to make up the 1,000 turns for the Yor\`{u}b\'{a} data.
The two Yor\`{u}b\'{a} sources are a mix of short dialogues in different scenarios including the market, home and school.
We call the collection of these corpora AfriWOZ.
It is interesting to note that though the data sizes are small, they are still larger than the COPA benchmark dataset available on the SuperGLUE \cite{wang2019superglue}.
In line with data acquisition standards \cite{bender2018data}, we provide the short data statement below and Table \ref{datachar} gives characteristics of the dataset.

\begin{quote}
    \textbf{\textit{Short data statement for the AfriWOZ  dataset.}}\\
    This is the AfriWOZ dataset for training and evaluating open-domain dialogue models.\\
    The licence for using this dataset comes under CC-BY 4.0.\\
    Total natural languages: 6 (Swahili, Wolof, Hausa, Nigerian Pidgin English, Kinyarwanda \& Yor\`{u}b\'{a})\\
    Total turns in the training set per language: 1,000\\
    Total turns in the validation set per language: 250\\
    Total turns in the test set per language: 250\\
    Domains covered in the data include hotel, restaurant, taxi and booking.\\
    The long version of this data statement is in the appendix.
\end{quote}

\begin{table}[h!]
\centering
\resizebox{\columnwidth}{!}{%
%\begin{tabular}{l|c|c}
\begin{tabular}{lcc}
\hline
\textbf{Language} &
\multicolumn{2}{c}{\textbf{Characteristics}}
\\
%\hline
 & Source & Translation method %& Tokens
\\
\hline
Pidgin English & M & HT %& 
\\
Yor\`{u}b\'{a}  & B+M & HT %& 
\\ % \hline
Hausa & M & MT+HR %& 
\\ % \hline
Wolof & M & HT %& 
\\
Swahili & M & HT %& 
\\
Kinyarwanda  & M & HT %& 
\\
\hline
\end{tabular}
}
\caption{\label{datachar} AfriWOZ dataset characteristics. Each contains 1,500 turns. \footnotesize{(M: MultiWOZ; B: Blog; HT: human translation; \acrshort{mt}: machine translation; HR: human review)}}
%}
\end{table}

\subsection{Translation Quality}
The translators, recruited online on Slack\footnote{slack.com/}, are native/L1 speakers of the target languages and second/L2 (but dominant) speakers of English.
They were to use either of the two possibilities for translation: human translation or \acrshort{mt} through Google \acrshort{mt} plus human review of all translations, for quality control (QC).
Each corpus is reviewed by the coordinator of each language.
Particularly, the Yor\`{u}b\'{a} language had a linguist review the data.
The risk of translating English conversations into unnatural conversations in the target languages was mitigated by using native speakers instead of just \acrshort{mt}.

\subsection{Translation Challenges}
The two main human translation challenges encountered include handling English entities and reframing English conversations for cultural relevance in the target languages.
Generally, entities in the data were retained, especially as this may facilitate \acrshort{mt} task.
In the future, this may be changed or two versions of the data maintained: one version with all the English entities and a second version with each language's common entities.
The experience and cultural background of the native speakers made it relatively simple to frame the English conversations into what seem natural in the target languages.

\section{Experiments}
\label{sec:method}
We compare 3 models: \acrfull{dialogpt} \cite{zhang2020dialogpt}, BlenderBot 90M \cite{roller-etal-2021-recipes} and a simple Seq2Seq with attention mechanism, as a baseline, based on the ParlAI platform by \citet{miller-etal-2017-parlai}.
Experiments  were  conducted  using a participatory approach \cite{nekoto-etal-2020-participatory} on Google Colaboratory with free GPUs.
Some experiments were on a shared DGX-1 machine  with 8 × 32GB Nvidia V100 GPUs.
The server  runs on  Ubuntu  18  and has 80 CPU cores.
Each experiment was conducted 3 times and the average perplexity (including standard deviation) was obtained.
%First, we discuss the data below before the models.

\subsection{Models}
The finetuning/training process for BlenderBot 90M and the seq2seq models was for about 20 minutes each.
Finetuning \acrshort{dialogpt} on each of the datasets for 3 epochs takes less than 20 minutes.
We did not do extensive hyperparameter search due to the constraints of time and resources.
The decoding algorithm across the models was set as top-k (k=100) and top-p (p=0.7).
We do not finetune/update the default tokenizer with new tokens or words from the target languages.
Instead, we leverage the default/generic tokenizers of the selected models.
We also attempted to compare the AfriBERTa tokenizer, which is trained for several African languages, by swapping it in for \acrshort{dialogpt}, but this was incompatible.
In addition, we recognize that the 3 models do not have exactly the same parameters or configuration and are, therefore, not expected to have the same performance.

\subsubsection{DialoGPT}
\citet{zhang2020dialogpt} introduced 3 sizes of the \acrshort{dialogpt}: the large, medium and small.
It is an English pretrained model for open-domain chatbots based on GPT-2.
It was trained on 147M turns of Reddit comments.
It uses \acrfull{bpe} tokenizer.
The medium model is reputed to have the best performance compared to its large and small versions.
In this work, however, we use the small version to minimize the problem of overfitting over small datasets.
We utilize the pretrained model from the HuggingFace hub and the generic autotokenizer \cite{wolf-etal-2020-transformers}.
The small model has 117M parameters, 12 layers and uses a vocabulary of 50,257 entries.
We use a batch size of 2 during finetuning because of memory constraints and perform ablation studies over the conversation context with values of 7 and 14, noting though that larger context sizes will bring memory challenges \cite{adiwardana2020towards}.

\subsubsection{BlenderBot 90M}

The model is a pretrained transformer model loaded from the ParlAI hub \cite{miller-etal-2017-parlai}.
It has 8 layers, 16 heads, uses Adam optimizer and byte-level \acrshort{bpe} for tokenization.
It has 87.5M trainable parameters, a batch size of 6 for finetuning and starts with the learning rate of 1e-5.
A variant of English Reddit discussions covering a vast range of topics and totaling 1.5B comments was used to train the model.
However, the data consists of group discussions instead of direct two-way conversational data.

\subsubsection{Seq2Seq}
The seq2seq is an encoder-decoder model that is based on the \acrshort{lstm} architecture \cite{hochreiter1997long} and uses the attention mechanism \cite{bahdanau2015neural}.
It was trained from scratch (random initialization) on the datasets in order to compare as a baseline.
The model has 805,994 trainable parameters and uses a batch size of 64.

\subsection{Evaluation}
For automatic evaluation, we follow \citet{adiwardana2020towards} and report only perplexity.
This is because automatic metrics typically used in \acrshort{mt}, such as \acrshort{bleu}, are poor for open-domain dialogue systems \cite{jurafsky2020speech,lundell2020conversational}.
Having multiple valid responses to prompts as reference is important for meaningful automated evaluations \cite{gangal-etal-2021-improving}.
These multiple valid responses are usually difficult to construct.
Probably the best evaluation is done by humans, though this may be subjective.
For human evaluation, we follow a similar method as in the original work by \citet{zhang2020dialogpt}.

\subsubsection{Perplexity}
Perplexity shows how well a model predicts a sample.
It minimizes the uncertainty of predicting the next token.
Ideally, the lower the perplexity, the better the model performs and the higher the perplexity, the more unsure the model is at predicting the next token \cite{adiwardana2020towards}.
This is used often to evaluate the language models built with n-grams of text dataset~\cite{sennrich2012perplexity}.
Perplexity has been shown to correlate with the human evaluation metric called Sensibleness and Specificity Average (SSA) by \citet{adiwardana2020towards}.
However, correlation of perplexity with human judgment is not always straightforward, as observed by \citet{roller-etal-2021-recipes} and \citet{hashimoto-etal-2019-unifying}.

\subsubsection{Human Evaluation}
We use the observer evaluation method, where evaluators (or annotators) read transcripts of conversation \cite{jurafsky2020speech}.
We ask human evaluators to rate single-turn conversations for human-likeness on a Likert scale with 3 entries (human-like (H), non-human-like (N) or uncertain (U)).
The reason is that lack of long-term contextual information is still an existing problem in conversational systems \cite{zhang2020dialogpt}.
A copy of each transcript is given to 3 native speakers per language to evaluate.
A total of 32 single-turn conversations are generated per language and 3 credibility test conversations spread out within the transcript (at positions 11, 21 and 26) to make up 35.
Putting more test conversations would have been desirable but we chose to balance this with the attention-span of the annotators, as lengthy transcripts demand more time.
A random list was generated and used to select the same 32 prompts for all the languages from each test set.
Only one model, which had the best perplexity across languages, was used to generate the conversations: DialoGPT c7 x 1,000 (having context size 7 and 1,000 training turns), though small scale human evaluation is carried out to verify sample conversations from the other models: BlenderBot 90M and the seq2seq.
The transcripts are available online$^2$.

Out of the total (24) transcripts returned, 6 were not credible.
Three credible evaluations per language were processed for result computation.
Simple majority vote decided the annotation of each single-turn conversation.
The credibility test conversations fulfil 2 goals: 1) they help us check if annotators are qualified or paying attention and 2) they introduce a new way to determine \acrshort{iaa} in a simple way, especially since the tests are homogeneous to the rest of the conversations.
We call this \acrfull{cus} and discuss it further in the next section.
Discredited evaluations are the ones that failed 2 or more out of the 3 credibility test conversations by marking them as anything but H.
The 3 credibility conversations are prompts and responses directly from the test set instead of generated responses from the model.
A simple instruction for every evaluator at the top of the transcript of conversations is given below.

\begin{quote}
    Below are 35 different conversations by 2 speakers. Please mark each one as Human-like (H) or Non human-like (N) or Uncertain (U) based on your own understanding of what is human-like.
\end{quote}

\textbf{Selection of evaluators}\\
The evaluators/annotators were recruited online on Slack.
They are also native/L1 speakers of the target languages and second/L2 (but dominant) speakers of English.
These are unbiased respondents who are not connected to the translation of the datasets nor did they take part in the training of the models.

\subsubsection{\acrshort{cus}}
%Raw percentages of observed agreement on a sample of annotated entities for measuring \acrshort{iaa} has been shown to be weak since some agreements may be due to chance \citep{clark2012handbook}, as discussed in Section~\ref{iaa}.
%Fleiss Kappa, another common \acrshort{iaa} metric, has been shown to be restrictive in its interpretation, depending on the number of categories \citep{landis1977measurement}, as Kappa is lower when the categories are more \cite{sim2005kappa}.
The basic assumption behind \acrshort{cus} is that if homogeneous samples that are introduced into the transcript can be used for establishing the credibility of the annotators, then they may be used for establishing their agreement.
It may be seen as a proxy over the entire transcript.
\acrshort{cus} is more intuitive, easier to calculate (as it's based on percentages) and seemingly less sensitive to changes in the number of categories being evaluated, compared to Fleiss Kappa (\textit{k}).
It is based on unanimous votes across the homogeneous samples.
The probability of obtaining high \acrshort{cus} rises when the benchmark score for annotator credibility is raised.
For example, if the benchmark scores for accepting annotators' work in two different jobs are 51\% and 71\%, then the probability of getting a higher \acrshort{cus} is higher in the latter.
This gives \acrshort{cus} an advantage over using raw percentages over the actual samples, due to the possibility of agreements by chance, which is likely in raw percentages.
%The homogeneous samples can be seen as a significant subset of the full transcript, especially when it fulfils the central limit theorem by having a minimum of 30 samples.

\section{Results \& Discussion}
\label{sec:res}

\subsection{Performance on African Languages}
Table \ref{models3} shows the perplexity results across the three models for the African languages.
\acrshort{dialogpt} with a context size of 14 achieves the best (lowest perplexity) result for each language, in the table.
This is inspite of using half the training size that is used for the BlenderBot 90M and Seq2Seq models.
Generally, \acrshort{dialogpt} performs best across the languages but there are languages that do not perform so well and the Hausa language Seq2Seq overfits.
In the relevant tables, sd, c7, and c14 stand for standard deviation, context size 7, and context size 14, respectively.
Also, bold figures are the better values per language.

\begin{table*}[h!]
\small
\centering
%\resizebox{\columnwidth}{!}{%
%\begin{tabular}{l|c|c}
\begin{tabular}{lcccc}
\hline
\textbf{Language} &
\textbf{Model} &
\textbf{Training turns} &
\multicolumn{2}{c}{\textbf{Perplexity}}
\\
%\hline
 & & & Dev (sd) & Test (sd)
\\
\hline
Nigerian Pidgin English & DialoGPT c14 & 500 & 67.57 (2.53) & 90.18 (3.24)
\\ % \hline
 & BlenderBot 90M & 1,000 & 81.23 (0) & 81.23 (0)
\\ % \hline
 & Seq2Seq & 1,000 & 277.2 (15) & 277.2 (15)
\\
\hline
Yor\`{u}b\'{a} & DialoGPT c14 & 500 & 12.63 (0.47) & 10.66 (0.40)
\\ % \hline
 & BlenderBot 90M & 1,000 & 154.43 (0.06) & 154.43 (0.06)
\\ % \hline
 & Seq2Seq & 1,000 & 45.85 (1.41) & 45.85 (1.41)
\\
\hline
Hausa & DialoGPT c14 & 500 & 26.40 (0.75) & 35.95 (0.73)
\\ % \hline
 & BlenderBot 90M & 1,000 & 39.39 (1.61) & 39.39 (1.61)
\\
 & Seq2Seq & 1,000 & 1.92 (0.12) & 1.92 (0.12)
\\
\hline
Wolof & DialoGPT c14 & 500 & 15.2 (0.09) & 26.41 (0.10)
\\ % \hline
 & BlenderBot 90M & 1,000 & 108.7 (0) & 108.7 (0)
\\
 & Seq2Seq & 1,000 & 401.6 (10.39) & 401.6 (10.39)
\\
\hline
Swahili & DialoGPT c14 & 500 & 20.03 (0.29) & 17.02 (0.22)
\\ % \hline
 & BlenderBot 90M & 1,000 & 128.8 (0.10) & 128.8 (0.10)
\\
 & Seq2Seq & 1,000 & 134.5 (2.75) & 134.5 (2.75)
\\
\hline
Kinyarwanda & DialoGPT c14 & 500 & 24.47 (0.17) & 26.45 (0.17)
\\ % \hline
 & BlenderBot 90M & 1,000 & 177.87 (0.06) & 177.87 (0.06)
\\
 & Seq2Seq & 1,000 & 195.07 (7.66) & 195.07 (7.66)
\\
\hline
\end{tabular}
%}
\caption{\label{models3} Results across the 3 main models}
\end{table*}

\begin{table*}[h!]
\small
\centering
%\resizebox{\columnwidth}{!}{%
%\begin{tabular}{l|c|c}
\begin{tabular}{lccc}
\hline
\textbf{Language} &
\textbf{Training turns} &
\multicolumn{2}{c}{\textbf{Perplexity}}
\\
%\hline
 &  & Dev (sd) & Test (sd)
\\
\hline
Nigerian Pidgin English & 500 & 42.55 (0) & 52.81 (0)
\\ % \hline
 & 1,000 & \textbf{37.95} (0.66) &  \textbf{46.56} (1.13)
\\
\hline
Yor\`{u}b\'{a} & 500 & 10.52 (0.04) & 9.65 (0.01)
\\ % \hline
 & 1,000 & \textbf{7.22} (0.06) &  \textbf{8.76} (0.08)
\\
\hline
Hausa & 500 & 18.53 (0.23) & 25.7 (0.4)
\\ % \hline
 & 1,000 & \textbf{9.92} (0.05) & \textbf{12.89} (0.04)
\\
\hline
Wolof & 500 & 15.2 (0.09) & 26.41 (0.10)
\\ % \hline
 & 1,000 & \textbf{14.91} (0.3) & \textbf{25.85} (0.04)
\\
\hline
Swahili & 500 & 15.55 (0.17) & 14.22 (0.14)
\\ % \hline
 & 1,000 & \textbf{9.63} (0) & \textbf{9.36} (0.03)
\\
\hline
Kinyarwanda & 500 & 19.28 (0.19) & 21.62 (0.22)
\\ % \hline
 & 1,000 & \textbf{10.85} (0) & \textbf{14.18} (0.08)
\\
\hline
\end{tabular}
%}
\caption{\label{abturns} Ablation study of \acrshort{dialogpt}-c7 over training turns. Bold figures are the better values per language.}
\end{table*}

\subsection{Performance vs. Amount of Data or Context size}
Taking the best model from Table \ref{models3}, which is \acrshort{dialogpt}, and doing ablation studies over the training set size and the context size, we obtain results in Tables \ref{abturns} and \ref{abcontext}, respectively.
We observe that increasing the training set size by doubling the number of turns brings improvement by lowering the perplexity for the model of each language.
Doubling the context size, however, does not have a similar effect.
Performance, in terms of perplexity, only improves when we half the context size from 14 to 7.
The better values are given in bold in each table.
The results are statistically significant, as all p-values (p < 0.0001) for the difference of two means of the two-sample t-test (between the two lowest results) for all the languages are smaller than the alpha (0.05).
Given that these results are obtained with small data, we believe increasing the data size will improve the results.

%It is probably not the best to compare perplexities across languages.
%The Nigerian Pidgin English (46.56) is higher than all the others while Yor\`{u}b\'{a} (8.76) has the lowest in test set results.

\begin{table*}[h]
\small
\centering
%\resizebox{\columnwidth}{!}{%
%\begin{tabular}{l|c|c}
\begin{tabular}{lccc}
\hline
\textbf{Language} &
\textbf{Context size} &
\multicolumn{2}{c}{\textbf{Perplexity}}
\\
%\hline
 &  & Dev (sd) & Test (sd)
\\
\hline
Nigerian Pidgin English & c7 & \textbf{37.95} (0.66) &  \textbf{46.56} (1.13)
\\
 & c14 & 70.21 (2.17) & 92.23 (2.33)
\\
\hline
Yor\`{u}b\'{a} & c7 & \textbf{7.22} (0.06) &  \textbf{8.76} (0.08)
\\
 & c14 & 7.63 (0.13) & 9.11 (0.14)
\\
\hline
Hausa &  c7 & \textbf{9.92} (0.05) & \textbf{12.89} (0.04)
\\
 & c14 & 11.30 (0.04) & 15.16 (0.05)
\\
\hline
Wolof & c7 & \textbf{14.91} (0.3) & \textbf{25.85} (0.04)
\\
 & c14 & 16.61 (0.2) & 30.37 (0.08)
\\
\hline
Swahili & c7 & \textbf{9.63} (0) & \textbf{9.36} (0.03)
\\
 & c14 & 11.07 (0.04) & 10.71 (0.05)
\\
\hline
Kinyarwanda & c7 & \textbf{10.85} (0) & \textbf{14.18} (0.08)
\\
 & c14 & 12.84 (0.1) & 17.43 (0.14)
\\
\hline
\end{tabular}
%}
\caption{\label{abcontext} Ablation study of \acrshort{dialogpt} over context sizes for training set with 1,000 turns. Bold figures are the better values per language.}
\end{table*}

\begin{table*}[h!]
\small
\centering
%\resizebox{\columnwidth}{!}{%
%\begin{tabular}{l|c|c}
\begin{tabular}{lcccccc}
\hline
\textbf{Model language} &
\multicolumn{4}{c}{\textbf{Scale (majority votes - 2/3)}} & \textbf{\acrshort{cus}} &
\textbf{Fliess \textit{k}}
\\
%\hline
 &  H (\%) & U (\%) & N (\%) & 3-way (\%) & \% &
\\
\hline
Nigerian Pidgin English & 78.1 & 0 & 6.3 & 15.6 & 66.7 & -0.079
\\
Yor\`{u}b\'{a} & 0 & 3.1 & 75 & 21.9 & 33.3 & -0.154
\\ % \hline
Hausa & 31.3 & 6.3 & 53.1 & 9.4 & 66.7 & 0.228
\\ % \hline
Wolof & 65.6 & 0 & 31.3 & 3.1 & 100 & 0.070
\\
Swahili & 28.1 & 15.6 & 34.4 & 21.9 & 66.7 & 0.067
\\
Kinyarwanda & 28.1 & 25 & 34.4 & 12.5 & 66.7 & 0.091
\\
\hline
 &
\multicolumn{4}{c}{\textbf{unanimous votes - 3/3}} & &
\\
\hline
Nigerian Pidgin English & 34.4 & 0 & 0 & - & 66.7 &
\\
Yor\`{u}b\'{a} & 0 & 0 & 25 & - & 33.3 &
\\ % \hline
Hausa & 12.5 & 0 & 21.9 & - & 66.7 &
\\
Wolof & 15.6 & 0 & 9.4 & - & 100 &
\\
Swahili & 9.4 & 0 & 9.4 & - & 66.7 &
\\ % \hline
Kinyarwanda & 9.4 & 0 & 6.3 & - & 66.7 &
\\
\hline
\end{tabular}
%}
\caption{\label{tab:humevalsummary}Human evaluation results of 3 annotators on 3 classes using single-turn conversations. A recent human-human upperbound is 92.1\%, according to \citet{adewumi2021sm}. \footnotesize{The subjective Kappa example of 2 annotators on 2 classes does not apply here since Kappa is lower when classes are more \cite{sim2005kappa}. - implies not applicable.}}
\end{table*}

\subsection{Human evaluation}
We observe from Table \ref{tab:humevalsummary} that the single-turn conversations of the Nigerian Pidgin English are judged as human-like 78.1\% of the time by majority votes.
34.4\% of them are unanimously judged as human-like, which is higher than both the 3-way split (when each annotator voted for each different category) of 15.6\% or non-human-like of 6.3\%.
This is intuitive, since Pidgin English is closely related to the English language, which is the language of pretraining.
Meanwhile, the Yor\`{u}b\'{a} transcript has 0\% human-like single-turn conversation.
%75\% non-human-like conversations, 3.1\% uncertain and 21.9\% are split 3-way. This makes Yor\`{u}b\'{a} the least language transferable to, among the set.
This may be because of a combination of reasons, including the language's morphology and written accent, among others.
It has the most peculiarities in written form, as shown in Table \ref{sampleconv}, making it challenging for the model.
%the relative quality of the dialogue data sources when compared to the MultiWOZ,
Wolof, Hausa, Kinyarwanda and Swahili follow after Nigerian Pidgin English with 65.6\%, 31.3\%, 28.1\% and 28.1\% of conversations judged as human-like, respectively.

\begin{figure}[h!]
\centering
\includegraphics[width=0.5\textwidth]{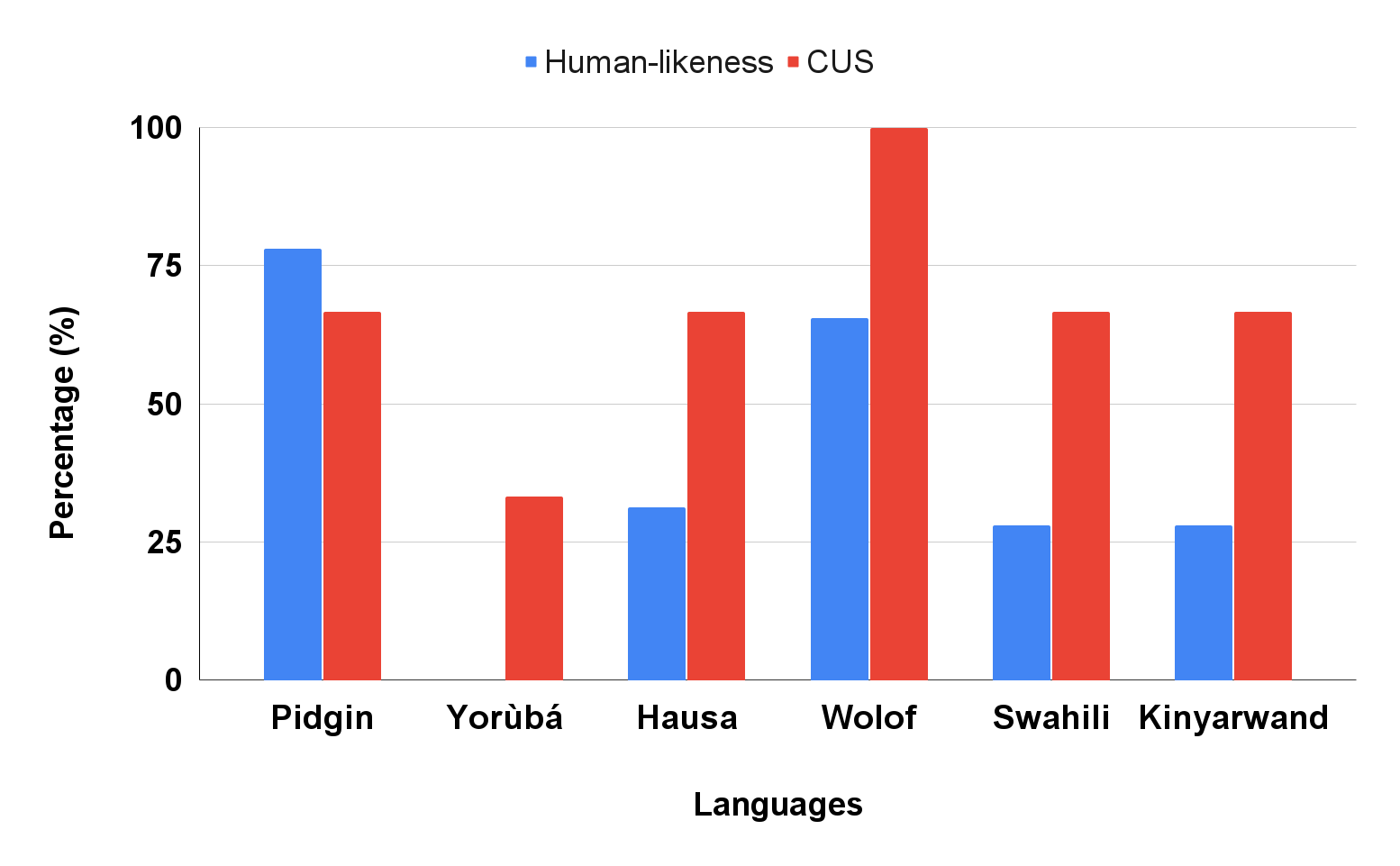}
\label{fig:chart}
\caption{Bar chart of human-likeness scores and credibility unanimous scores (cus) for the languages. Generally, \acrshort{cus} is high for the languages, except for Yor\`{u}b\'{a}.}
\end{figure}

Figure \ref{fig:chart} depicts the human-likeness scores and the credibility unanimous scores for the languages, as given in Table \ref{tab:humevalsummary}.
When we compare the best-performing (Nigerian Pidgin English) with a recent human-human upper-bound (92.1\%) for conversations, given by \citet{adewumi2021sm}, we observe that this best-performing model is still behind in terms of performance.

%The least credibility \acrshort{iaa} unanimous score for any of the languages is 33.3\%.
%This is the score for Yoruba while Pidgin English, Kinyarwanda, Swahili and Hausa achieve 66.7\% and Wolof achieves 100\%.

The Fleiss \textit{k} scores are not interpretable using the Kappa 2 annotators on 2 classes guide \cite{landis1977measurement}, as our study uses 3 annotators on 3 classes and the Kappa is lower when the classes are more \cite{sim2005kappa}.
Indeed, our study confirms the observation made by \citet{gwet2014handbook} that the interpretation guide may be more harmful than helpful.
\citet{perez-almendros-etal-2020-dont} also report how the Kappa score rose from 41\% to 61\% when the classes were reduced from 3 to 2. 
We believe that \acrshort{cus} is a more intuitive and reliable method, at least, in this case because of the homogeneous test conversations.

\subsection{Error Analysis}
%Some of the challenges with open domain conversational systems are well-known \cite{roller-etal-2021-recipes, zhang2020dialogpt}.
We highlight a few of the challenges of the models.
\acrshort{dialogpt}, though the best model in our investigation, shows weaknesses, particularly with the Yor\`{u}b\'{a} language by failing to generate the correct word morphology, sometimes.
Below are 2 examples where the conversations are not coherent.

\begin{itemize}
    \item \textit{prompt:} \d{E}s\d{\`{e}} \'{n} ro m\'{i}
    \item \textit{response:} E\`{e}l\'{o}\`{o} ni or\'{u}k\d{o} \d{o}d\'{u}n?
    \item \textit{prompt:} \`{I}y\d{e}n n\'{a}\`{a} da, w\d{\'{o}}n \d{s}\`{e} \'{n} dar\'{u}gb\'{o} n\'{a}\`{a} ni
    \item \textit{response:} K\`{i} ni or\'{u}k\d{\'{o}} or\'{u}nju r\d{E}?

\end{itemize}

On the other hand also, BlenderBot 90M mixes English responses with its generation sometimes, even though the prompts are of the foreign target language.
Two examples of prompts in Hausa and responses are given below.
There, we also see that it echoes the prompt given to it in the first pair of example.

\begin{itemize}
    \item \textit{prompt:} Barka dai, me za ku iya gaya mani game da gidan abincin bangkok na burni?
    \item \textit{response:} barka dai , za za ku iya gaya mani game da gidan abincin bangkok na burni ? i ' m not sure if you ' re being sarcastic or not .
    \item \textit{prompt:} Ee na mutum 2 da 19:30 na ranar Lahadi don Allah. Ina bukatan lambar nuni kuma.
    \item \textit{response:} haha , i ' m not sure if i ' m the only one who thinks that .
\end{itemize}

A major drawback of the seq2seq model is the overly repetitive nature of some of its responses across different prompts.
Two examples of the prompts and generated responses for the Nigerian Pidgin English are given below.
We see exactly the same response for two different prompts and the response is not coherent.

\begin{itemize}
    \item \textit{prompt:} I need one place to chop for the center wey dey expensive. Good bye
    \item \textit{response:} I dey go like .
    \item \textit{prompt:} e dey okay. i fit get that number. you fit also recommend me hotel wey dey expensive
    \item \textit{response:} I dey go like .
\end{itemize}

\subsection{Post-Evaluation Feedback}
Some of the evaluators, in post-evaluation feedback, explained that coherence of the conversation mattered as a deciding factor in their judgment.
So did the grammar.
For example, considering Yor\`{u}b\'{a}, responses that referenced inanimate objects as if in the context of animate objects or humans were voted as non-human-like.
Wrong placement of an accent mark or character was also motivation for voting conversations as non-human-like.
For Wolof, many of the conversations are human-like except for cases where the responses were inconsistent with the prompt or question given.
For example,  there were conversations that were hard to judge because the responses are questions to the prompts, which happen to be questions themselves.
Such conversations were awarded the uncertain votes by the particular annotator.

\section{Related Work}
\label{sec:related}
There are a number of pretrained models for open-domain conversational systems.
Some of them include Texar \cite{hu2018texar}, DLGnet \cite{olabiyi2019multiturn},  Meena \cite{adiwardana2020towards} and BlenderBot \cite{roller-etal-2021-recipes}.
These are pretrained on dialogue datasets.
In BlenderBot 2 \cite{komeili2021internet,xu2021beyond}, the same BlendedSkillTalk (BST) \cite{smith-etal-2020-put} collection of datasets  used for BlenderBot 1 \cite{roller-etal-2021-recipes} is used to train the model, in addition to 3 others.
There exist, also, models pretrained on large text and adapted for conversational systems.
Such models include T5 \cite{JMLR:v21:20-074} and BART \cite{lewis-etal-2020-bart}.
Another pretrained model on conversational data, \acrshort{dialogpt}, was trained on Reddit conversations of 147M exchanges \cite{zhang2020dialogpt}.
In single-turn conversations, it achieved performance close to that of humans in open-domain dialogues.
DialoGPT is based on GPT-2 \cite{radford2019language}.
It is an autoregressive model, which achieved \acrshort{sota} results in different \acrshort{nlp} tasks \cite{radford2019language}.

\citet{solaimanprocess} observed different harmful outputs in GPT-3, the successor of the GPT-2 model.
They discovered that a mitigating factor is carefully curating a small dataset, which determines the behaviour of the model outputs.
They made a good case for fine-tuning non-toxic text compared to reducing toxicity through controllable methods using filters or control tokens.
Topics such as history, science and government were covered in the dataset \citep{solaimanprocess}.
The 80 texts in the values-targeted dataset utilized by \citet{solaimanprocess} range in length from 40 to 340 words.

Recently, \citet{artetxe-etal-2020-cross} hypothesised that deep monolingual models learn some abstractions that generalise across languages, while working on cross-lingual transferability.
This is in contrast to the past hypothesis that attributes the generalization ability of multilingual models to the shared subword vocabulary used across the languages and joint training, as demonstrated for m\acrshort{bert} \cite{pires-etal-2019-multilingual}.
Besides m\acrshort{bert}, there other multilingual deep models \cite{ogueji-etal-2021-small,devlin2018mbert,NEURIPS2019_c04c19c2,reid-etal-2021-afromt,xue-etal-2021-mt5}.
The performance of such multilingual models on low-resource languages and unseen languages are known to be relatively poor \cite{pfeiffer2020AdapterHub, wang2021towards}.

In evaluating the performance of open-domain chatbots, it has been shown that automatic metrics, like the BLEU score, can be very poor but they are still used in some cases \cite{lundell2020conversational}.
Conversation turns per session is another metric of interest
\cite{zhou2020design}.
Perplexity is also widely used for intrinsic evaluation of language models and its theoretical minimum, which is its best value, is 1 \cite{adiwardana2020towards}.
Probably the best evaluation is done by human evaluators (or annotators) but this can be subjective.
The judgment of human evaluators is seen as very important, especially since humans are usually the end-users of such systems \cite{zhang2020dialogpt}.

\section{Conclusion}
\label{sec:conc}
In this study, we presented the new high-quality AfriWOZ dataset for dialogue modelling for 6 African languages.
We also demonstrated the cross-lingual transferability hypothesis for the 6 African languages and observe that it is possible to different degrees of success.
The English pretrained \acrshort{dialogpt} model resulted in the best perplexity scores across the languages.
%and provided us the reason to generate single-turn dialogues for human evaluation.
%Nigerian Pidgin English appears to have the most transferable properties.
%and has the best human evaluation results.
%The hypothesis that deep monolingual models learn some abstractions that generalize across languages appears to hold.
AfriWOZ may be extended to the total 143,000 dialogue turns in the MultiWOZ to achieve better performance in modelling.
Better performance may also be achieved if the tokenizers are optimized on the target languages by training from scratch or finetuning, as this will allow more native tokens to be represented.
It may be worthwhile to construct a transferability index/matrix for various languages.
This will indicate the amount of benefit that may be harnessed from utilising such properties in different downstream tasks.
%Having shown that the cross-lingual transferability hypothesis of deep monolingual models appears to hold, at least for English as a pretraining/source language, is it also possible that other/target languages have such capabilities in a reverse training?
%This may be an interesting query to investigate.

\section{Limitations}
\label{sec:limits}
The data used to finetune the models are relatively small and cover only a few domains, hence, the generation capabilities of the models are limited.
Furthermore, though we made effort to use carefully curated dialogue data and avoid \acrfull{pii}, the potential to generate offensive output is still present, as the pretrained models retain biases in the pretraining data.

%\section{Ethics Statement}

\section*{Acknowledgment}
Our profound appreciation goes to Angela Fan, Clemencia Siro and Wilhelmina Onyothi Nekoto for their contributions to this work. Also, we wish to thank the many evaluators who judged the conversation transcripts for the various languages.
%and the anonymous reviewers of this paper for their valuable feedback.

\bibliography{tacl2021}
\bibliographystyle{acl_natbib}

\iftaclpubformat

\onecolumn

\appendix

\begin{table}[h]
\centering
%\resizebox{\columnwidth}{!}{%
\begin{tabular}{p{.17\textwidth}|p{.8\textwidth}}
\multicolumn{2}{l}{Data statement for the AfriWOZ dataset for open-domain dialogue \& other \acrshort{nlp} models.}
\\
 \hline
\textbf{} &
\textbf{Details}
\\
\hline
Curation rationale & Due to the unavailability of dialogue data for low-resource African languages, this dataset was created.
\\
\hline
Dataset language & Swahili, Wolof, Hausa, Nigerian Pidgin English, Kinyarwanda \& Yor\`{u}b\'{a}
\\
\hline
 & \textbf{Demographics of contributors}
 \\
\hline
No of contributors & 19
\\
\hline
Age & -
\\
\hline
Gender & Male \& Female
\\
\hline
Language & L1
\\
\hline
 & \textbf{Demographics of annotators}
\\
\hline
No of annotators & Not applicable
\\
\hline
 & \textbf{Data characteristics}
\\
\hline
Total samples & 1,500 turns per language
\\
\hline
Total natural languages & 6 (Swahili, Wolof, Hausa, Nigerian Pidgin English, Kinyarwanda \& Yor\`{u}b\'{a})
\\
\hline
Training set turns per language & 1,000
 \\
\hline
Validation set turns per language & 250
 \\
\hline
Test set turns per language & 250
\\
\hline
Domains covered & hotel, restaurant, taxi and booking.
\\
\hline
Base data & \acrshort{multiwoz} and 2 blogs for Yor\`{u}b\'{a} only.
\\
\hline
 & \textbf{Others}
\\
\hline
\acrshort{iaa} & \acrshort{cus} 33.3\% - 100\%
\\
\hline
Licence & CC-BY 4.0.
\\
\hline
\end{tabular}
%}
\caption{\label{aCappsvanalogy}}
\end{table}

\printglossary[type=\acronymtype]
\end{document}